# An Adiabatic Capacitive Artificial Neuron With RRAM-Based Threshold Detection for Energy-Efficient Neuromorphic Computing

Sachin Maheshwari, *Member, IEEE*, Alexander Serb, *Senior Member, IEEE*, Christos Papavassiliou, *Senior Member, IEEE*, and Themistoklis Prodromakis, *Senior Member, IEEE*

*Abstract*—In the quest for low power, bio-inspired computation both memristive and memcapacitive-based Artificial Neural Networks (ANN) have been the subjects of increasing focus for hardware implementation of neuromorphic computing. One step further, regenerative capacitive neural networks, which call for the use of adiabatic computing, offer a tantalising route towards even lower energy consumption, especially when combined with 'memimpedace' elements. Here, we present an artificial neuron featuring adiabatic synapse capacitors to produce membrane potentials for the somas of neurons; the latter implemented via dynamic latched comparators augmented with Resistive Random-Access Memory (RRAM) devices. Our initial 4-bit adiabatic capacitive neuron proof-of-concept example shows 90% synaptic energy saving. At 4 synapses/soma we already witness an overall 35% energy reduction. Furthermore, the impact of process and temperature on the 4-bit adiabatic synapse shows a maximum energy variation of 30% at $100^oC$ across the corners without any functionality loss. Finally, the efficacy of our adiabatic approach to ANN is tested for 512 & 1024 synapse/neuron for worst and best case synapse loading conditions and variable equalising capacitance's quantifying the expected trade-off between equalisation capacitance and range of optimal power-clock frequencies vs. loading (i.e. the percentage of active synapses).

*Index Terms*—Adiabatic, artificial neural networks, energy-efficient, memristor, neuromorphic computing, RRAM.

## I. Introduction and Motivation

NEUROMORPHIC computing, coined by Carver Mead in late eighties [1], and more generally 'brain-inspired computing' has emerged in recent years as a key direction of

Manuscript received December 10, 2021; revised May 15, 2022; accepted June 6, 2022. This work was supported in part by the Engineering and Physical Sciences Research Council (EPSRC) Programme Grant Functional Oxides for Reconfigurable Technologies (FORTE) under Grant EP/R024642/1 and in part by the Royal Academy of Engineering (RAEng) Chair in Emerging Technologies under Grant CiET1819/2/93. This article was recommended by Associate Editor M.-F. Chang. *(Corresponding author: Sachin Maheshwari.)*

Sachin Maheshwari, Alexander Serb, and Themistoklis Prodromakis are with the Centre for Electronics Frontiers, School of Engineering, The University of Edinburgh, Edinburgh EH9 3FB, U.K. (e-mail: maheshwari.sachin@ed.ac.uk; aserb@ed.ac.uk; t.prodromakis@ed.ac.uk).

Christos Papavassiliou is with the Department of Electrical and Electronic Engineering, Imperial College London, London SW7 2AZ, U.K. (e-mail: c.papavas@imperial.ac.uk).

Color versions of one or more figures in this article are available at https://doi.org/10.1109/TCSI.2022.3182577.

Digital Object Identifier 10.1109/TCSI.2022.3182577

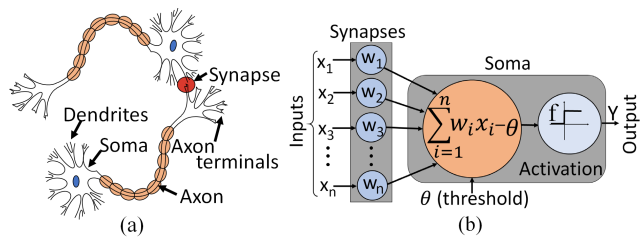

Fig. 1. (a) Biological neurons connected by Synapses. Soma = neuron body. (b) Typical artificial neuron (AN) block diagram where synapses are modelled as programmable weights and the soma represents the summation and the activation function.

future electronics design. It is intended to mimic the function of biological neural systems with the aim of approaching their compact size/weight and low power consumption while solving the Von-Neumann bottleneck [2] through parallelism. The need for this approach becomes ever more accentuated as increasingly large amounts of data are used both when training and using the large neural networks of today.

The most fundamental components of neurons, be they biological or artificial, are generally the synapses and the somas, as shown in Fig. 1, with synapses typically outnumbering neurons by large factors (256:1 for neuromorphic chips such as ROLLS [3], 10k:1 for neocortical pyramidal cells [4] and even 250k:1 for cerebellar Purkinje cells [5]). In hardware, this translates to vast numbers of power-hungry multipliers performing the basic operation of a synapse: multiplying synaptic inputs with synaptic (weights).

Over the years, many neuromorphic synapse implementations have been developed including switch capacitor [6], sub-threshold FET [7], Li-ion synaptic transistor [8], non-multiplier synapse [9] and oxide-RAM [10]. In addition, capacitive-based synapses [11] have also been explored and deployed in neural networks to improve space-energy efficiency in comparison to MOS devices. Yet, the energy-efficient hardware implementation of these learning systems is particularly challenging due to their intensive computation, memory, and communication that is necessary for real-time learning and classification.

In order to mitigate this problem, several solutions have been proposed, with a particularly promising strand of research in the direction of single-component synapses. For example, there has been substantial work in memristive synapses [12]–[14] where a single, tuneable resistor acts as a weight by exploiting Ohm's law (an input voltage is naturally multiplied by a tuneable resistance - or a current with a conductance). Memristive devices have shown great improvement in terms of dense integration property and as well as low energy consumption.

However, resistive computing is dissipative, hence ideas have emerged on building tuneable memcapacitors [15], [16]. Some applications of memcapacitors include memcapacitive synapse with integrate and fire neurons [16], memcapacitive crossbars [17], [18] and logic applications [19], [20]. This perspective has recently become much more attractive due to its lower power consumption horizon, better emulation of neural activities [16] and zero sneak current leakage issue [17]. And yet, even capacitive computing relies on a shuttling charge from a power supply to the ground. In this endeavour, the next logical step in reducing net energy dissipation is to investigate adiabatic computing, whereby charge sent into capacitors partially returns to the power supply periodically.

Adiabatic Logic [21] (AL) is a low power technique. Unlike pure CMOS logic that works using a fixed DC supply, AL operates with an AC power supply (the 'power clock' - PC) that changes gradually from zero to a maximum allowed voltage and back, returning a fraction of the energy used to power the 'upswing' or 'evaluation' phase during the 'downswing' or 'recovery' phase. Additionally, energy dissipation depends on the speed of the evaluation and recovery phases: faster changes mean larger currents, which in turn imply bigger losses due to the inevitable parasitic resistive elements of the circuit. Theoretically, for an infinitely small slow-changing power signal, the adiabatic dissipation reaches zero [22]. The PC is practically designed using either inductors or capacitors, which enables energy recycling during the downswing/discharge phase [23]- [25]. Though it has been in existence for more than three decades, showing energy-efficient operations [26], [27], it failed to create an impact in the field of reversible computing mainly due to the complex multi-phase power-clock generator design. In the field of neuromorphic engineering, one could exploit the benefits of the charge recovery property of adiabatic logic in the design and development of adiabatic artificial capacitive synapse banks.

The operating principle of the regenerative capacitive synapse has been shown in [29] with basic analysis and energy comparison simulation results versus non-regenerative (non-adiabatic) synapses. In this paper, we substantially extend the analysis, provide mathematical approximations for our fundamental findings, include the comparator/soma in our calculations, perform key parametric analyses indicating how the synaptic loading, temperature and basic process corners affect optimal running frequency and energy dissipation and thus demonstrate a basic, but functionally complete artificial neuron building block circuit that uses capacitive synapses and a RRAM-based threshold detection circuit in depth. Additionally, here we use pMOS body biasing to reduce the energy consumption of the capacitive synapses (an upgrade over [29]) and compare the results with the non-adiabatic design.

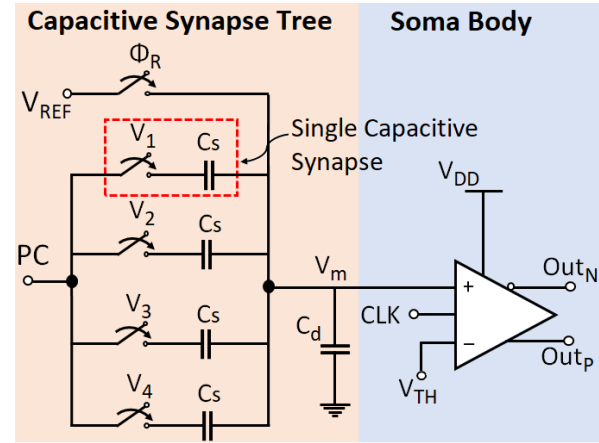

Fig. 2. Schematic of the proposed adiabatic capacitive artificial neuron (ACAN): The power-clock '$PC$' is generated from a chip-wide common power-clock generator, the identical capacitive synapses, $C_S$ (representing different synapses) activated via switches $V_1, V_2, V_3, V_4$ (corresponds to pre-neuron inputs), generating 16-input combinations. Baseline voltage, $V_{REF}$ is applied during the reset operation, controlled by $\phi_R$. The *damping capacitor*, $C_d$ limits the maximum voltage swing achievable on the *membrane potential* node ($V_m$). Finally, the DLC comparator (acting as an activation function) generates a binary output based on the difference between the membrane potential ($V_m$) and the firing threshold ($V_{TH}$).

The paper is structured as follows: Section II outlines the system architecture and operational principle. This section demonstrates the overall working of the proposed Adiabatic Capacitive Artificial Neuron ($ACAN$) which is followed by an in-depth explanation of the individual logic blocks. Section III demonstrates the simulation results using SPICE for a commercially available $0.18 \mu$m CMOS technology for various frequencies, scaling projection for increased convergence ratio and corner analysis. Challenges and opportunities for future work are discussed in Section IV. Finally, section V, concludes the paper.

## II. SYSTEM ARCHITECTURE AND BASIC OPERATION

The standard Artificial Neuron (AN) implements a simple equation - sum of input-weight products followed by thresholding and an 'activation function' (1):

$$Y = f\left[\left(\sum_{i=1}^{n} w_i x_i\right) - \theta\right] \quad (1)$$

where, $Y$ is the output, $w_i$ are the synaptic weights, $x_i$ the inputs, $\theta$ is the threshold value or bias, $n$ is the fan-in (no. of synapses) to the AN, $f$ is the activation function and $i$ simply indexes over all input lines of the AN ($i = 1, \cdots, n$).

Our proposed ACAN embodiment of (1) used in this work consists of two distinct blocks as shown in Fig. 2: a) adiabatic capacitive synaptic tree. b) cell soma body. The synaptic tree is powered using a specially designed power-clock ($PC$), orchestrating the recycling of charge. Whereas, a fixed DC voltage ($V_{DD}$) powers the soma design using a memristive (resistive random access memory - RRAM) Dynamic Latched Clocked Comparator (DLCC).

The ACAN is an AN: With reference to Fig. 2, each $V_i$ is the corresponding input $x_i$ to the AN that represents pre-neuron inputs. If the switch $V_i$ is ON ($x_i$ is active), then the up-swinging 'voltage wave' is allowed to propagate from the PC to the corresponding synaptic capacitor $C_s$. The capacitive synapse are identical with each represents a different synaptic connection. The modulation of the pre-neuron inputs with its synaptic capacitance generates a *membrane voltage*, $V_m$. The membrane potential serves as one of the input to the $DLCC$ (soma body serves as an activation function) which makes a decision based on the comparison to the *threshold value*, $V_{TH}$ (firing threshold) of the $DLCC$. When $V_m$ exceeds $V_{TH}$, the output comparator (Soma) generates a binary value of 1. Overall, the illustrated unit acts as a standard, 4-synapse McCulloch-Pitts neuron [31] with Heaviside step activation function and binary input/output values $x_i$, Out ∈ {0, 1}, which is a popular choice, although other activation functions also exist [32].

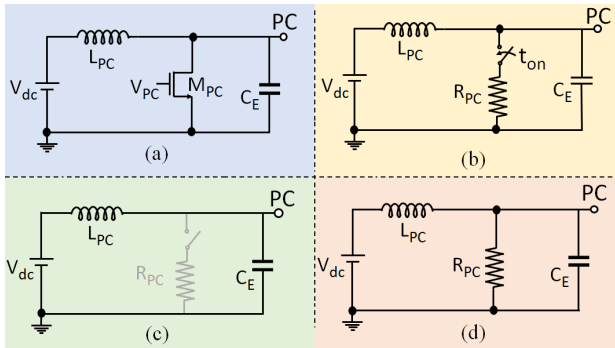

Fig. 3. (a) A single-phase resonant power clock generator with one inductor ($L_{PC}$), by-pass nMOS switch ($M_{PC}$) and an equalising capacitor ($C_E$). (b) Equivalent RLC model with ideal switch and nMOS ON-resistance ($R_{PC}$). (c) When transistor ($M_{PC}$) is OFF, the circuit is a pure LC oscillator (barring parasitics). (d) When the transistor ($M_{PC}$) is on, the circuit decomposes into RLC and dissipates energy through ($R_{PC}$).

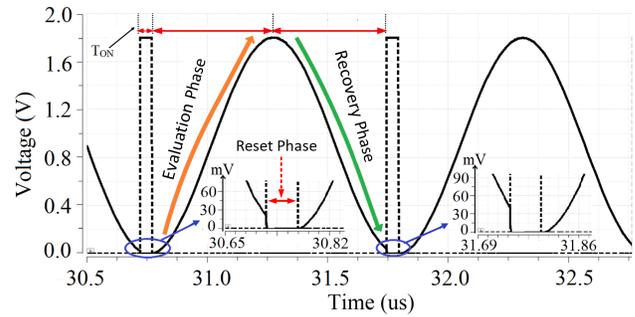

Fig. 4. Typical power-clock waveform, taken at $1MHz$. Dotted lines denote the nMOS by-pass switch turning on for a brief time interval $t_{ON}$ to top-up the energy in the system (reset phase shown in inset). Note how at the end of each recovery phase the PC voltage doesn't quite reach zero again (insets). The vertical drop in voltage indicates energy dumping from the capacitor to ground. Had the recovery phase been allowed to continue until the time derivative of the PC voltage zeroes the PC voltage would still not have reached zero, reflecting energy losses due to non-idealities in the LC. The by-pass transistor is tasked with replenishing these losses. In this example we used $L_{PC} = 1mH$, $C_E = 25pF$ and a minimum size $M_{PC}$ transistor.

### A. Powering the Adiabatic Capacitive Neurons

Adiabatic logic works on two basic principles for energy minimization [21]; i) Only open the switch when there is no potential difference across it and ii) Only close the switch when no current is flowing through it. Adiabatic logic requires a specialized power supply scheme, typically sinusoidally [23] varying power rails in order to enable computation, although trapezoidal [33] and stepwise charging [25] power-clocks also exist. Stepwise Charging Circuit (SCC) power-rails can be energy-efficient, especially if we use more steps [24], however, an increased number of steps tends to increase energy cost again due to the circuit overheads associated with the control signals managing the corresponding capacitors [27].

In our case we are using a sinusoidal, LC-based power clock with a by-pass nMOS switch, $M_{PC}$ entraining the LC circuit formed by the combination of our PC inductor, $L_{PC}$, equalising capacitor $C_E$ and load capacitance $C_L$ (itself a combination of parasitics + the capacitance's of active synapses) and compensating for natural energy losses in the system. This is shown in Fig. 3, along with it's equivalent ideal lossless model. Where $R_{PC}$ is the ON resistance of $M_{PC}$. The desired oscillating frequency determines the values of the inductor and equalising capacitor:

$$L_{PC}C_L = \frac{1}{4\pi^2 f_{PC}^2} = \frac{T_{PC}^2}{4\pi^2} \quad (2)$$

where $f_{PC}$ is the oscillating frequency and $T_{PC}$ is the power-clock period.

The practical waveform of a single-phase sinusoidal $PC$ is shown in Fig. 4. The PC behaves as a (lossy) free LC oscillator during the *Evaluation* and *Recovery* phases, interrupted by a top-up phase, *reset Phase*. A large fraction of the operating energy thus oscillates between the capacitor and the inductor, topped-up in brief bursts of duration $t_{ON}$ by the by-pass transistor.

The voltage time-evolution trace in Fig. 4 suggests loss modelling via the classical RLC circuit equation:

$$v(t) = v_{max} \cdot Re\left\{e^{j(\omega t+\theta)-\lambda t}\right\} \quad (3)$$

where $v_{max}$ is the lossless voltage magnitude which is ideally equal to $V_{dc}$, $\omega = 2\pi/f_{PC}$ is the angular frequency, $\theta$ is the phase angle which depends on the initial conditions and $\lambda \geq 0$ represents losses. However, we note that $\lambda$ assumes 'smooth' loss profile (as would be expected by the presence of resistive elements in the circuit) - in practice, additional factors such as 'spillover' caused by accidentally moving outside the power supply ranges, forward-biasing over/under-voltage protection structures and non-linear leakages may introduce additional non-linear losses, complicating the loss profile picture.

By-pass transistor ($M_{PC}$) is tasked with replenishing all these losses as efficiently as possible: too much top-up and the system reaches equilibrium via increased losses due to e.g. spillover; too little and the system may end up oscillating at an excessively low amplitude (e.g. where the 'approximately-$\lambda$' losses are exactly replenished by the PC). In exchange, a burst of energy dissipation occurs when $M_{PC}$ is turned ON (see reference model in Fig. 3(d). $M_{PC}$ turns ON only for a brief period $t_{ON} = D \cdot T_{PC}$ in each clock period to compensate for the incurred energy losses, where $D$ is the duty cycle.

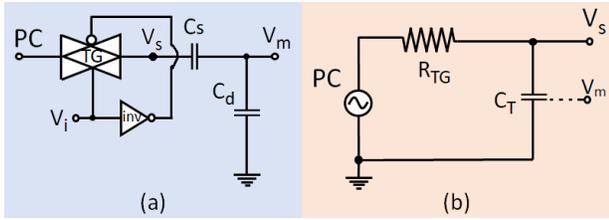

Fig. 5. (a) A basic, single-synapse tree showing the transmission gate ($TG$) switch and the minimum size inverter for generating the complementary control signal. (b) Equivalent RC loading model of a single capacitive synapse. Where, $R_{TG}$ is the resistance of the $TG$ and $C_T$ is the equivalent capacitance which is a series combination of $C_s$ and $C_d$ plus the aggregate parasitic capacitance, $C_{par}$.

Everything else being equal, the energy dissipation increases with higher $M_{PC}$ width (lower ON resistance) and $D$.

Next, we note that losses are load-dependent. The width of $M_{PC}$ can be set to meet the requirements of a fixed, baseline load, whilst tweaking $D$ can then compensate for (possibly dynamic) load variation above/below baseline. The external equalising capacitor ($C_E$) helps moderate the effects of load variation (arising due to capacitance differences in each cycle). However, higher $C_E$ values tend to increase the energy dissipation as more charges shuttle through the LC resonator and hence associated resistive losses inflict more dissipation. The load variation phenomenon can be seen in Fig. 4. In the first power-clock cycle, the minimum voltage, $V_x$ reached before $M_{PC}$ is turned ON is around 25 mV whereas, in the next cycle the value is close to 45 mV. The energy dissipated when $M_{PC}$ switch is turned ON is given by:

$$E_{PCG} = \frac{1}{2} C_{PC} V_x^2 \left(1 - e^{\frac{-t_{ON}}{R_{PC} C_{PC}}}\right) \quad (4)$$

We note that: a) Typically low $R_{PC}$ means that the $C_{PC}$ (total capacitance at node $PC$ i.e $C_E + C_{parasitic}$) ≈1. b) In the case where the PC over-tops up the lost energy it is possible that at the end of the cycle the PC voltage is below GND. In this case the energy is still lost as we enforce a current down a dissipative element (impedance is real). c) for the sake of simplicity in the subsequent sections, the PC system will supply the synaptic trees, but not the cell somas. Because the synaptic trees are purely capacitive networks, they present no DC path to the ground except leakages due to parasitic.

### B. Adiabatic Capacitive Synaptic Tree

The conceptual cornerstone of the adiabatic capacitive synaptic tree is using the power clock to periodically bias a capacitive divider structure (see Fig. 2), where each participating capacitor $C_s$ is gated (it is allowed to charge if the gate is open and not allowed otherwise). The resultant voltage appears at the common node of the divider, which we call by analogy to biology the 'membrane potential' $V_m$ of the neuron. Fig. 5a illustrates the simplest possible case: a single-synapse tree with its transmission gate switch.

The damping capacitor $C_d$ transmits voltage changes at the input to $V_m$ via capacitive voltage division:

$$\Delta V_m = \frac{C_s}{C_s + C_d} \Delta V_{PC} \quad (5)$$

We now walk through the full operating cycle for the generalised system: Beginning with the *Reset Phase*, the baseline-voltage ($V_{REF}$) is forced upon the *membrane potential* via $\Phi_R$ (see Fig. 2). Note that the input terminals are implicitly zeroed as the PC is at its trough at this stage.

Next, at the start of the *Operation Cycle*, $\Phi_R$ is turned OFF and the input switches, connecting the synaptic capacitance's ($C_s$), are configured ON or OFF as necessary, ready for the next upswing of the PC's *evaluation* phase. In this phase, the top plate of activated $C_s$ caps start charging and follow the PC. At the PC peak, $V_m$ charges to a voltage level $> V_{REF}$ as per the n-synapse capacitive divider equation:

$$V_m = V_{pk} \frac{\sum_i (x_i \cdot C_{si})}{\sum_i C_{si} + C_d + C_{par}} + V_{REF} \quad (6)$$

where $V_{pk}$ is the peak PC voltage (where sampling by the soma's DLCC takes place), $x_i$ is the binary input signal to the synaptic TGs that corresponds to switch $V_i$, $C_{si}$ denotes the $i$th synaptic capacitance, $C_d$ denotes the damping capacitance and $C_{par}$ represents the combined effective parasitic to ground. The equation includes the characteristic "sum of products" equation that defines the behaviour of a classical McCulloch-Pitts artificial neuron, with the capacitance of each $C_s$ acting as a weight.

Finally, the *recovery* phase begins, the PC starts falling, the activated synaptic capacitance's ($C_s$) discharges and the current flows back to the PC.

Overall, we note that the "baseline" reference voltage $V_{REF}$ actively defines the lowest voltage that $V_m$ is allowed to take during normal operation. Simultaneously, the extra capacitance $C_d$ acts as a damper on the voltage swing range on $V_m$, helping accommodate the input swing of the DLCC. For example: suppose the comparator swings between 0.5 V and 1.5 V under a 2 V power supply. In this case, we would set $V_{REF} = 0.5$ V (the lower limit) and our maximum swing equal to $V_{DD}/2$, which implies that $C_d \approx \sum_i C_{si}$.

An important aspect of the design concerns the bulk connection of the synapse transistors. The threshold voltage ($V_t$) is a function of body-to-source voltage ($V_{BS}$), reducing when forward biasing the body terminal [34]. Meanwhile, we want a lower threshold during the complete 'power-clock' cycle. This can be achieved by connecting the TG's pMOS bulk to the PC, as per Fig. 6. In the evaluation phase the bulk bias is weaker because the bulk is not linked to $V_{DD}$, resulting in a threshold drop. Whereas, in the recovery phase the PC has dropped so much that the bulk diode forward biases. This way we ensure 2 things: 1) In the evaluation mode we retain reverse bias on the bulk, so the corresponding Cs charges only if the TG is active. 2) In the recovery mode we use the forward bias of the bulk connection to more efficiently empty the charge from Cs. The energy comparison between the bulk-to-VDD & bulk-to-PC is demonstrated in Table I for a single capacitive

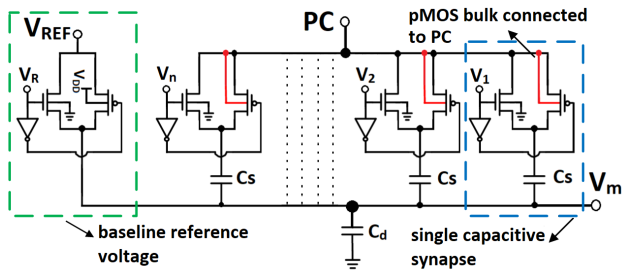

Fig. 6. Transistor level diagram of n-bit capacitive synapse. Each consisting of nMOS & pMOS transistor, an inverter and a capacitor. The pMOS bulk (shown in red) is connected to the PC aiding in energy saving. $C_S$ activated via switches ($V_1, V_2, \ldots, V_n$). Baseline voltage, $V_{REF}$ is applied during the reset operation, controlled by $V_R$. $C_d$ acting as a *damping capacitor*, limits the maximum voltage swing achievable on the *membrane potential* node ($V_m$).

TABLE I
ENERGY DISSIPATION PER OPERATION OF A SINGLE SYNAPSE VS. TG PMOS BULK CONNECTION APPROACH. CIRCUIT UNDER TEST AS PER (FIG. 5a) WITH $C_s = 1$ pF AND $C_d = 1$ pF

| Power Clock Frequency | Synapse Energy (fJ) | | Saving |
|---|---|---|---|
| $f_{PC}$ | $to-PC$ | $to-V_{DD}$ | % |
| 100KHz | 6.46 | 7.59 | 14.89 |
| 500KHz | 36.10 | 41.80 | 13.64 |
| 1MHz | 65.80 | 76.70 | 14.21 |
| 10MHz | 598.30 | 674.50 | 11.30 |

synapse at different frequencies. It is this design that is benchmarked in section III.

We now analyse the energy dissipation of a single capacitive synapse. The dissipation consists of two components; 1) charging and discharging of the inverter used to drive the pMOS transistor of the TG, 2) adiabatic charging and discharging of the synapse capacitance. Modelling Fig. 5a as a simple RC circuit powered by the sinusoidal power supply (as shown in Fig. 5b). The total energy dissipation $E_S$ [1] is given by;

$$E_S = C_T V_{DD}^2 \cdot \frac{\pi^2}{8} \cdot \frac{R_{TG} C_T}{T_{PC}} + C_{inv} \cdot V_{DD}^2 \quad (7)$$

where the 1st term denotes the adiabatic loss and follows the formula derived in [35] (loss for RC circuit under sinusoidal stimulation) and the 2nd term is the standard inverter energy dissipation per switching event. $C_{inv}$ is the equivalent parasitic input capacitance at the gate terminal of the pMOS transistor in the TG switch. Naturally, parasitics will cause some deviation from this ideal.

C. RRAM-Based Threshold Detection

The membrane voltage generated by the synaptic tree in II-B needs to be sensed using sensing circuitry. The choice of the comparator was mainly providing zero static power dissipation. In literature, MRAM-based sensing circuits have been well developed featuring low-power and high speed [36], [37]. Here, we have incorporated our in-house developed RRAM device [38] to demonstrates its advantage in the conventional Dynamic Latched Clocked Comparator (DLCC) by providing tunability.

[1]Does not include the non-adiabatic energy dissipation of the synapses.

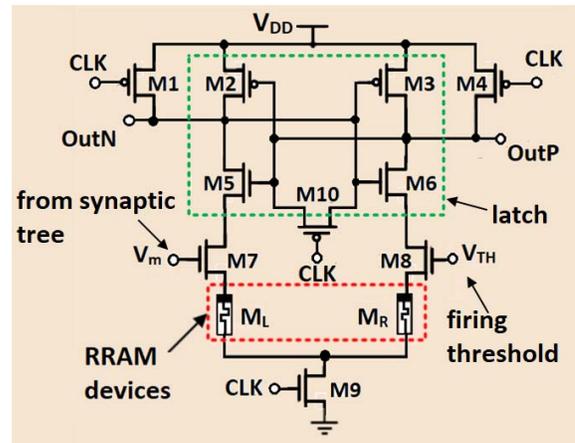

Fig. 7. Transistor level diagram of a dynamic latched clocked comparator (DLCC) with RRAM devices ($M_L$ & $M_R$). It generates a binary output corresponding to the difference between $V_m$ and the $V_{TH}$ reference input (spike or no spike). Note the DLCC (neuron soma) is powered by a regular supply $V_{DD}$. The sizes of all the transistors are set to technology minimum.

With reference to Fig. 7, DLCC is a strong-arm comparator with tuneable source degeneration at the input differential pair. The DLCC is augmented with RRAM devices, $M_L$ and $M_R$, that tune its offset to accommodate systematic imperfections. When the clock ($CLK$) signal is low, the output nodes are pre-charged to $V_{DD}$ and the DC path to the ground is cut-off. The comparison process starts when $CLK$ goes high: During the comparison phase, initially the outputs are disconnected from $V_{DD}$ and switching current source $M9$ begins to conduct and pulls down the bottom node of the RRAM device and invariably the source node of $M7$ and $M8$. The initial currents through $M7$ and $M8$ depend now both on their gate voltage levels and the degeneration, which can favour one transistor or the other in the critical initial phase of positive feedback. When $V_m > V_{TH}$, the gate of $M3$ ends up pulled down before $M2$ has a chance to do so and the latch triggers to output node $OutN$ pulled to ground and $OutP$ pulled up to $V_{DD}$.

An oversized $M_L$ (or correspondingly $M_R$) can decrease the Vgs across $M7$ sufficiently to overcome $V_m > V_{TH}$ (up to a point), as shown in Table II. In practice, the system is intended to operate after the memristive devices have been set appropriately. For instance: to tune the system at 50:50 ratio of the $DLCC$ output, $V_m$ is set to a desirable threshold voltage (by appropriately manipulating $V_{REF}$) and then tuning the memristive devices (via trial and error).

Fig. 8 shows the overall DLCC switching functionality. We run 400 cycles (1 $\mu$s/cycle for a total duration of 400 $\mu$s) and sweep input $V_m$ across a full range of 1.8 V (200 $\mu$s ascending and 200 $\mu$s descending) with threshold ($V_{TH}$) set to 1.1 V and RRAM devices set to 10 $k\Omega$ each (based on model from [38], resistance values quoted under standard read-out voltage of 0.2 V). Here the resistive state and the dynamicity depends on the bias voltage and on the time-step of the transient pulse respectively. However, the resistive state is independent of the other physical parameters like temperature and process and for which in-situ calibration for the $DLCC$ offset need to be done, especially if the desired value is not zero. A behavioral model for temperature dependence of nonvolatile switching dynamics of $TiO_x$ memristors is presented in [39].

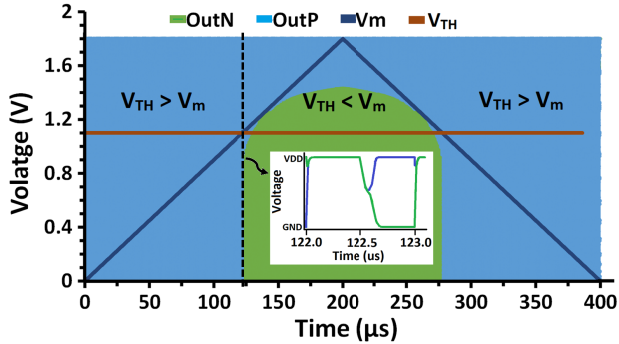

Fig. 8. DLCC basic functionality test: Input $V_m$ (dark blue trace) is slowly swept between [0 V → 1.8 V → 0 V] over 400 $\mu$s and input $V_{TH}$ (red trace) remains stable at 1.1 V throughout. Because this run contains 400 decisions where OutP and OutN repeatedly reset to $V_{DD}$ and then settle to either $V_{DD}$ or $GND$ -see inset-, the transient output traces merge into a blur. To read the final outcome of each decision we inspect the colour of the (always single) trace that successfully reaches 0 V in any given cycle. This is green between approx. 122 − 277 $\mu$s, indicating a state of [OutP=1, OutN=0], and blue elsewhere. The simulation was ran for balanced RRAM devices ($M_L = M_R = 10\,k\Omega$). The results (nominal device parameters, no noise) indicate offset below the resolution of this technique and no hysteresis.

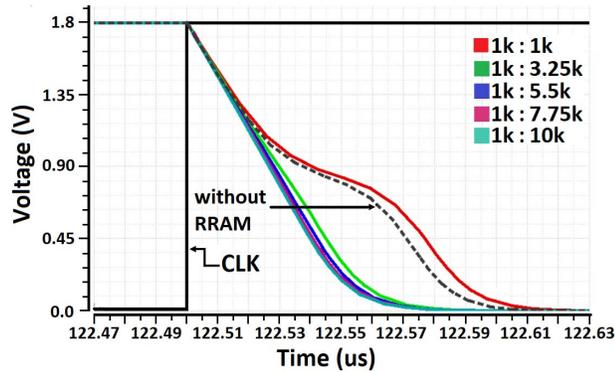

Fig. 9. Delay tuning test for the RRAM device. The black dotted trace is for DLCC without RRAM devices. RRAM device $M_L$ is fixed at 1 $k\Omega$ whereas, $M_R$ is swept from 1 $k\Omega$ to 10 $k\Omega$. The balance RRAM device (red trace) shows maximum delay, more than the dotted black trace. By tuning to a higher resistive state the delay can be tuned to a lower value.

Next, we check the DLCC's time-to-decision when $M_R, M_L \in 1, 10\,k\Omega$. The worst-case falling delay clocks in at ≈87 ns for the case where $M_L=M_R=1\,k\Omega$ as shown in Fig. 9. We note that because we operate in the 'clinically clean' environment of no noise and nominal devices these metastability-induced delays are expected to be upper bounds. Our worst-case falling delay indicates that the $DLCC$ can reach a solid decision while the PC is "approximately static at its maximum peak", even for differential inputs that are extremely close (within $X\,mV$) of the true offset. For completeness of the result, we plot delay versus $V_m$ for 4 corner values and 1 without RRAM devices as shown in Fig. 10. However, for $M_L:M_R$ ratio of $10\,k\Omega:1\,k\Omega$ there is no crossover and thus it has been omitted. The maximum delay was observed for high resistance balanced ratio value (10 $k\Omega$:10 $k\Omega$), whereas the least delay (lower than without RRAM) was measured for unbalanced ratio value of 1 $k\Omega$:10 $k\Omega$. All the delays were measured when the output reaches 10 % of $V_{DD}$.

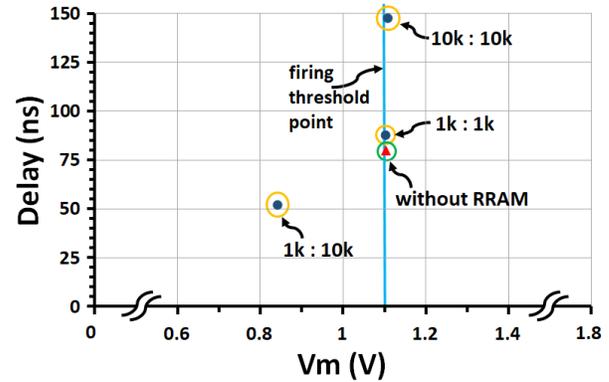

Fig. 10. Delay vs input voltage plot for 3 corner values (blue dot) and 1 without RRAM devices (red triangle). for $M_L : M_R$ ratio of 10 $k\Omega$: 1 $k\Omega$ there is no crossover hence it has been omitted. The delay for the DLCC without RRAM devices (red triangle) is ≈79 ns. The maximum delay of ≈147 ns is measured for a balanced condition with high resistance values (10 $k\Omega$) whereas, minimum delay of ≈51 ns is measured for $M_L:M_R$ of 1 $k\Omega$:10 $k\Omega$.

TABLE II
OFFSET VOLTAGE OF DLCC IN $mV$ VS. RRAM DEVICE RESISTIVE STATES. THE SIMULATION IS RUN FOR 1000 CYCLES (1 $\mu$s/CYCLE FOR A TOTAL DURATION OF 1000 $\mu$s) AND SWEEP INPUT $V_m$ ACROSS A FULL RANGE OF 1.8 V (500 $\mu$s ASCENDING AND 500 $\mu$s DESCENDING)

| $M_L \setminus M_R$ | 1 $k\Omega$ | 3.25 $k\Omega$ | 5.5 $k\Omega$ | 7.75 $k\Omega$ | 10 $k\Omega$ |
|---|---|---|---|---|---|
| 1 $k\Omega$ | 0.20 | 110.0 | 178.4 | 225.2 | 261.2 |
| 3.25 $k\Omega$ | -154.6 | 0.19 | 90.2 | 153.2 | 196.4 |
| 5.5 $k\Omega$ | -343.6 | -116.8 | 0.18 | 77.6 | 131.6 |
| 7.75 $k\Omega$ | -674.8 | -233.8 | -91.6 | 0.25 | 66.8 |
| 10 $k\Omega$ | no crossover | -397.6 | -190.6 | -77.2 | 0.3 |

Finally, we repeat the basic functionality experiment from Fig. 8 for $M_L, M_R$ independently swept between [1, 10] $k\Omega$ and obtain a table of offsets vs. $M_L/M_R$ (table II). We observe that; 1) The overall circuit has a wide trimming range spanning from <-$V_{DD}$/2 (no-crossover detected) to ≈+261 $mV$. 2) The maximum offset occurs at maximum $M_L$. 3) The table is asymmetric, indicating that common mode voltage affects the result.

The presented DLCC is connected to the synaptic tree, with $V_m$ receiving the weighted sum voltage obtained from the capacitive synapses. The intended operation is to trigger the DLCC when the PC is at its peak, obtain a decision before leaving the immediate peak and then remain latched or reset until the next peak. This is naturally achieved by synchronising the DLCC's regular clock with the PC as will be shown in Fig. 11. It can be easily shown that the behaviour of the DLCC with capacitive synaptic tree input is described by:

$$OutP = V_{DD} \cdot stp \left\{ V_{pk} \left( \frac{\sum_i x_i \cdot C_{si}}{C} \right) + V_{REF} - (V_{TH} - V_{os}) \right\} \quad (8)$$

where $stp$ stands for the Heaviside step function (version with range 0, 1) and $V_{os}$ is the offset voltage resulting from the presence of $M_L$ and $M_R$. $C$ is the total capacitance given by the denominator of eq. (6).

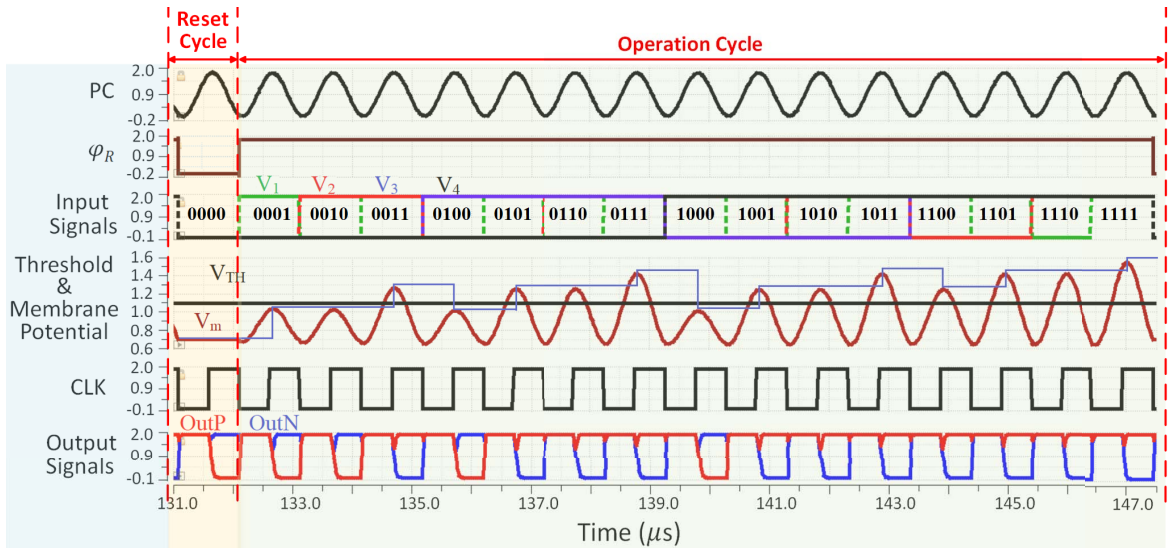

Fig. 11. Timing waveform for a 4-bit ACAN at $1\,MHz$ CLK frequency. 1st trace: power-clock waveform. 2nd trace: Reset/Operation mode control signal. The baseline voltage ($V_{BL}$) is refreshed in the reset cycle. 3rd trace: Standard, 4-bit, binary-weighted input configuration generating all possible 16-states. 4th trace: Membrane potential $V_m$ evolution. Because in this example all synapse weights are the same, we can split all inputs into 5 equivalence classes based on the number of 1's they feature. The peak magnitude of $V_m$ voltage (tracked by the blue trace) depends on the equivalence class and maximizes when all the TG switches are turned ON (all 1's). 5th trace: DLCC (non-adiabatic) clock (CLK). 6th trace: Binary output of the DLCC. Note that when $V_m$ exceeds $V_{TH}$, $OutN$ goes low (blue trace dips) and when $V_m$ is below $V_{TH}$, $OutP$ dips low (red trace).

The DLCC acts as a neural "soma" with an individualised firing threshold ($V_{TH} - V_{os}$) consisting of a (globally adjustable) "baseline" $V_{TH}$ plus a significant capacity of local, non-volatile adjustment thanks to the locally integrated RRAM-based memory. This is evident by directly comparing eqs. (1) and (8). A similar but detailed analysis of the RRAM based detection circuitry is provided in [40].

## III. PERFORMANCE BENCH-MARKING

In this section we investigate the performance of a complete, 4-synapse ACAN with particular focus on the synaptic tree and the balance of energy dissipation between synaptic tree and soma. This focus reflects the large synapse/soma ratio in typical neural networks (e.g. 256:1 in ROLLS chip [3], 10 k:1 in the actual human cortex [4]).

Performance evaluation of the proposed ACAN is done using Spectre simulator in Cadence EDA tool for a $0.18\,\mu m$ commercially available CMOS technology. All synaptic weights are kept equal for simplicity. We use $V_{DD} = 1.8\,V$ synaptic tree baseline $V_{REF} = 0.7\,V$ and the neural firing threshold $V_{TH} = 1.1\,V$. As previously, devices $M_L$ & $M_R$ are modelled using our own memristor model [38] with resistive states of $10\,k\Omega$ for both. The load capacitance of the DLCC output nodes is fixed at $1\,pF$.

### A. Base Scenario and Set-up

Initial functional verification and energy estimate extraction is performed on a "base scenario" configuration of the system. The tuneable parameters are set as follows: $t_{ON} = 50\,ns$ and power supply capacitance $C_E = 25\,pF$ and inductance $L_{PC} = 1\,mH$. CLK frequency was $1\,MHz$.

TABLE III

AVERAGE AND EXTREME MAX AND MIN 'membrane potential' ($V_m$) VALUES (IN V) FOR ALL INPUT EQUIVALENCE CLASSES (SAME NUMBER OF ACTIVE INPUTS) ACROSS ALL 5 RUNS (1 UNSCRAMBLED AND 4 SCRAMBLED DATA SET). THE MAXIMUM DIFFERENCE WAS OBSERVED ACROSS THE 5 SETS FOR EACH EQUIVALENT CLASS

| Equivalence class | Average (V) | | Extremes (V) | | Max difference (mV) |
|---|---|---|---|---|---|
| | max | min | max | min | |
| 1-bit active | 1.046 | 1.035 | 1.050 | 1.012 | 38 |
| 2-bit active | 1.277 | 1.264 | 1.283 | 1.270 | 35 |
| 3-bit active | 1.438 | 1.435 | 1.444 | 1.417 | 27 |
| 4-bit active* | 1.563 | – | 1.570 | 1.544 | 26 |

\* For a 4-bit synapse only one combination will have 4-bit active, hence there is only one single average value.

The baseline circuit is subjected to a transient analysis whereby the inputs are swept across all binary codes. Next, the binary coded input is randomly scrambled four times and the run is repeated. The response of the DLCC is monitored, with $V_m$ also traced for diagnostic reasons. The timing waveform for the unscrambled run is shown in Fig. 11. Input/output (I/O) mapping consistency is verified by recording the peak voltage levels for all input equivalence classes (same number of active inputs) across all runs. Maximum peak voltage discrepancies for each equivalence class are shown in Table III. The monotonicity of the I/O mapping is preserved and the digital output is completely consistent across all runs.

The variations in $V_m$ voltage peaks across each equivalence class arise as a result of the PC behaviour's history-dependence; itself a result of the load-dependence of losses in the PC (see section II-A). Tightening control over the conditions at the start of each evaluation phase is an involved subject and requires development of an adaptive controller for the reset process (e.g. parameter $t_{ON}$ in Fig. 4); a technique reminiscent of "maximum-power point tracking" systems in solar power cells [41]. This will be the subject of future work.

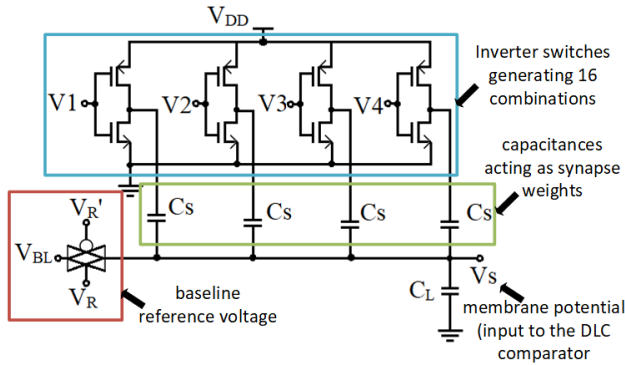

Fig. 12. Non-adiabatic capacitive 4-synapse neuron. Minimum size CMOS inverters are used ($Wn_{min} = Wp_{min} = 220\,nm$, $Ln_{min} = Lp_{min} = 180\,nm$). This design undergoes the same set of inputs as our proposed ACAN.

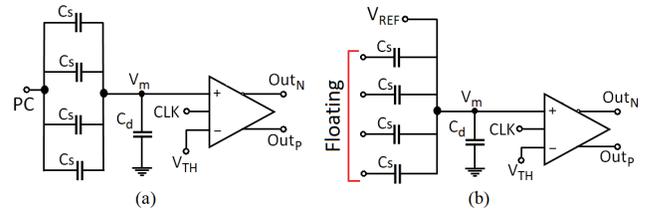

Fig. 13. Illustration of 4-synapse neuron circuit when (a) fully-loaded, i.e. all the synapses are engaged/ON (b) minimally loaded, i.e. all synapses are disengaged/OFF. Note: it is always in the later state that we take the opportunity to reset $V_m$ to $V_{REF}$.

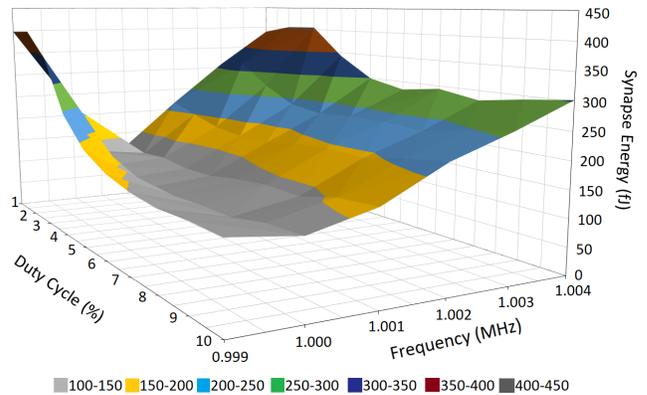

Fig. 14. Energy dissipation/cycle of the capacitive synapse tree for the case where all inputs are OFF (no capacitors engaged - value is worst average energy/operation taken across a window of 20 cycles length; see text for more information). The graph covers 6x frequencies close to $1\,MHz$ and duty cycle ($D$) ranging from 1% to 10%. All other parameters as baseline scenario.

A notable detail in Table III is the special treatment of the case where 0 bits are active. In this case we take the opportunity to force a reset of the synaptic tree baseline voltage in order to guard against gradual drift induced by any leakage currents (also see Fig. 13b). Thus, the all-0 input case is not technically a valid input to the DLCC, but will always reliably lead to a no-fire output anyway. In practice and with neurons possessing larger synaptic trees the opportunity to reset the baseline voltage may arise so infrequently that a separate mechanism for forcing a period re-calibration may need to be used (e.g. force global re-calibration every X clock cycles).The design and operation of such module is outside the scope of the present work.

For energy dissipation estimation we track energy consumption for every cycle with non-zero input, then average it across all runs and all input equivalence classes to obtain an estimate of operating energy performance under an input signal with uniform statistics. For the base scenario this figure is $4.68\,pJ$, of which $4.49\,pJ$ is dissipated by the soma at $1\,MHz$.

### B. Parametric Analysis and Bench-Marking

A number of key factors affect the performance of the ACAN. These include: a) the frequency of operation (influenced by the PC's LC components, $C_E$, $L_{PC}$), b) the set-up of the top-up phase duration ($t_{ON}$), c) the geometry of transistor $M_{PC}$ enforcing the top-up phase within the PC (focus on width $W_n$) and d) the loading (how many synapses draw current from the PC at any given cycle). All of these interact in non-trivial ways, but here we will start elucidating some of the key design trade-offs and show their practical effects. For final bench-marking/comparison purposes we use a non-adiabatic counterpart to our 4-synapse ACAN as shown Fig. 12. The DLCC is the same in both designs.

*1) Operating Timings: Frequency and Reset Duty Cycle:* To begin, we investigate how the different operation timings affect energy-efficiency. To do this we look at two key parameters: power clock frequency and duty cycle of the reset phase. For this experiment we were interested in finding the optimal operating conditions yielding the lowest energy for a *fixed load*. As such we took the extreme signals of all-0 and all-1, as shown in Fig. 13. Next, we note that even with fixed input signal (i.e. capacitive loading) not every cycle is the same: the initial state of the evaluation phase will depend on the end state of the preceding recovery phase, for example the magnitude of the 'jump' shown in the inset of Fig. 4. With sufficiently long operation under fixed load some kind of equilibrium may be reached, be it a fixed point or some limit cycle. In order to account for this effect we ran a total of 600 cycles (from $t = 0$ to $t = 600\mu s$), removed the initial 200 cycles when the system is still recovering from start-up transients and then operated a sliding window of length 20 cycles across the remaining 400 samples. We then took the average energy per operation within each 20-cycle window and kept the *worst value* obtained. This was done to ensure that we both have a worst-case value and at the same time eliminate any potential 'outlier' values arising from any putative limit cycle behaviour. The overall results of this effort are shown in Figs. 14 and 15.

The input signal cases depicted in Figs. 14 and 15 denote extreme cases where the neuron would remain either completely silent for a protracted period of time or receive synaptic spikes at every synapse in every cycle. More general inputs are expected to perform optimally somewhere between these extremes. For the present, however, we make the following observations: First, the optimal operating frequency between the extreme input cases varies by about 4%. This figure will depend on the ratio between the synaptic capacitors under consideration and the power clock's internal cap (here a capacitance of 4 pF in the synapses vs. a 25 pF in the power clock yielding a ratio of 16% - this drops to approx. 8% if we consider the equivalent capacitance seen by the PC, including $C_L$). Second, we note that the minimum energy dissipation's differ quite substantially from a minimum of $\approx$104 $fJ$ at 1 MHz and 5% duty in the all-0 case to $\approx$206 $fJ$ at 963 $kHz$ and 2% duty in the all-1 case, with the more heavily loaded case clearly consuming more energy. Third, the sensitivity of the energy function with duty cycle is perhaps surprisingly limited, with only moderate effects on energy dissipation caused by an order-of-magnitude change in duty cycle. Fourth, the sensitivity with respect to operating frequency is very steep, with energy doubling for a 2‰ change away from optimum. Fifth, with less capacitive load, the sensitivity vs. frequency increases, possibly indicating that other energy dissipation factors such as spillover start dominating very quickly. Finally, duty cycle and frequency effects are not always independent as can be seen by the fact that the optimum energy 'valley' in the all-0 load graph is not parallel to the duty cycle axis. This finer effect requires dedicated investigation.

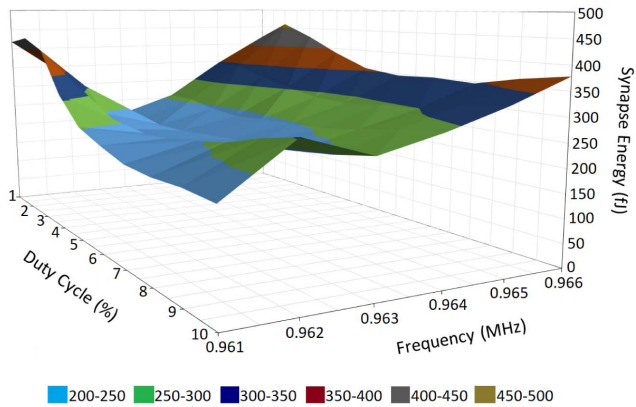

Fig. 15. Same as Fig. 14, but for the case where all inputs are ON (all capacitors engaged). The shift in optimum frequency away from the nominal 1MHz value is very clearly visible.

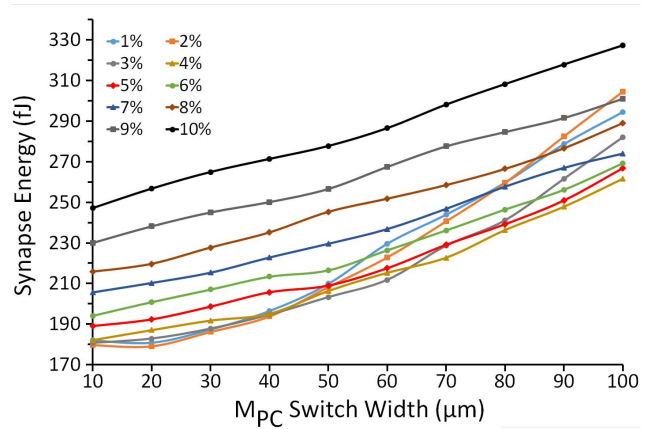

Fig. 16. Worst-case energy dissipation/cycle per capacitive synapse vs. reset phase duty cycle and PC transistor width (average over a run consisting of 16 unique inputs, 1x input/cycle - worst run out of 5 shown). Operating frequency is 977kHz. Line colours denote duty cycle.

*2) Power Clock Transistor Width and Duty Cycle:* Next, we investigate the interplay between the width of the power clock transistor and the duty cycle of the reset phase. For this test we used the same approach for assessing energy dissipation as for the baseline scenario: 1× unscrambled and 4× randomly scrambled binary code sweeps were created and ran. Each sweep was ran 32 times yielding a total run time of 32× runs times 16× cycles (1× cycle/possible input), i.e. 512 $\mu$ s. Then, in a spirit similar to the frequency-duty cycle analysis, we intended to take the run clocking the worst energy dissipation among the last 5x runs across all scrambled/unscrambled data-sets as the worst-case performance estimate. However, we found out that when comparing the energy dissipation across the last 5× runs of each sweep they were all within 0.1% of each other. Hence we used the worst-case *sweep average* energy dissipation (worst of the 5× scrambled/unscrambled sweeps) to generate each data-point in Fig. 16 for each unique combination of PC transistor width and duty cycle. In this test, the base frequency was set to 977kHz, i.e. approximately 2% below the no-load case of 1 $MHz$ and roughly in the middle of the frequencies identified for no-load and full-load cases of Figs. 14 and 15, as one might expect on a loading figure on average at 50% of maximum *for such a large ratio between equalising synaptic capacitance* -see section III-D for more on that subject-. This was confirmed to be very close to optimum for the given input. All other parameters were as in the baseline scenario. Upon this basis we then varied duty cycle $t_{ON}$ between 1-10% and the width of the PC transistor ($M_{PC}$) $W_n$ between 10-100 $\mu m$.

Results indicate an expected monotonic increase in energy/operation as the width of $M_{PC}$ increases, raising both the ability of the PC generator to draw current via the inductor uninhibited by voltage drops across its parasitic resistance and the load capacitance. However, the dependence of energy on duty cycle is far less straightforward: at high $W_n$ and lower duty cycle we notice a U-shaped dependency with a minimum around the 4-6% mark. At low $W_n$ it appears that this optimum is reached at around the 1% mark, though it is unclear whether the energy dissipation would increase if we dropped duty cycle below our minimum quantum of 1%. Seen from another perspective, for duty cycles between 1-4% energy dissipation vs. $W_n$ increases roughly quadratically whilst from 5% and above the increase is linear. It is still unclear what causes this precise behaviour but the working hypothesis is that: a) at "low" duty cycles the system behaves as usually expected (the $t_{on}$ phase replenishes losses in the LC resonator), but because the replenishing rate is too high the swing of the PC increases causing both "over-spill" and increased losses to parasitic resistances due to the higher voltages prevailing. This is corroborated by PC swings of 2.2V+ recorded at 1% duty and 80+ $\mu m$ transistor width. b) at "high" duty cycles the system works self-correctively, i.e. if the system has excess energy and PC swings below GND (i.e. the opposite of what happens in the insets of Fig. 4), the transistor will have time to remove some of that energy, reducing over-spill and increased parasitic losses. Indeed results show that for duty cycle $\geq$ 3% the PC swing never exceeds $1.9V$ for all data points in Fig. 16. c) as $W_n$ decreases the excess energy replenishing the LC resonator decreases, thus mitigating the PC swing increase and shifting the optimum operating duty cycle closer to 1%. The effect of power clock disturbance at 5% duty cycle for variable width is shown in Fig. 17. We note that for the particular cycle (all one input left side and all zero input right side) shown in the figure, the power clock "steals" energy from the resonator in the case of $W_n = 100\mu m$ (red trace is below ground when the reset phase starts).

TABLE IV
COMPARISON OF THE ENERGY DISSIPATION/CYCLE OF A 4 SYNAPSE/NEURON BETWEEN ADIABATIC AND NON-ADIABATIC IMPLEMENTATIONS FOR A RANGE OF HIGHLY DIFFERENT SCENARIOS. WHILST COMPLETE PARAMETER OPTIMISATION HAS NOT BEEN DONE FOR EACH INDIVIDUAL CASE RESULTS SHOW THAT: A) SIGNIFICANTLY LOWER SYNAPTIC ENERGY DISSIPATION IS ACHIEVABLE IN A WIDE RANGE OF SCENARIOS AND B) SIMILAR POWER FIGURES ARE ACHIEVED IN THE RANGE FROM 100kHz TO 1MHz. TRANSISTOR CHANNEL LENGTHS MINIMUM IN ALL CASES

| Nominal Frequency ($f_{PC}$) | Power-Clock Parameters | | | | Energy/Clock-Cycle (pJ) | | | | | |
|---|---|---|---|---|---|---|---|---|---|---|
| | | | | | ACAN | | | Non-Adiabatic | | |
| | $W_n$ ($\mu m$) | $t_{ON}$ (ns) | $C_E$ (pF) | $L_{PC}$ (mH) | Synapses | DLCC | Total | Synapses | DLCC | Total |
| 100KHz | 200 | 500 | 100 | 25 | 0.246 | 4.499 | 4.744 | 2.651 | 4.600 | 7.251 |
| 500KHz | 100 | 100 | 50 | 2 | 0.186 | 4.491 | 4.677 | 2.651 | 4.601 | 7.252 |
| 1MHz | 30 | 50 | 25 | 1 | 0.189 | 4.495 | 4.684 | 2.653 | 4.601 | 7.254 |
| 10MHz | 50 | 5 | 25 | 0.010 | 0.904 | 4.884 | 5.788 | 2.651 | 4.964 | 7.615 |

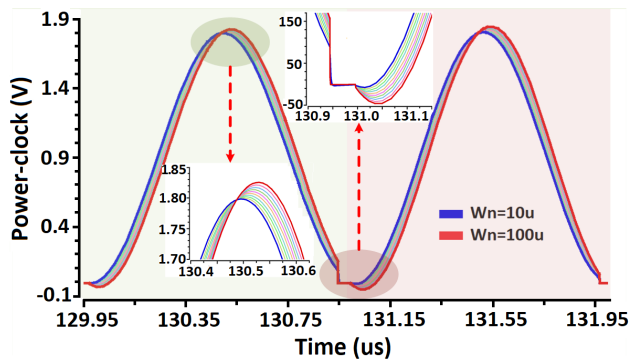

Fig. 17. Two cycles of the power-clock for $W_n$ changing from $10\mu$m to $100\mu$m at 1 $MHz$ frequency and 5% duty cycle ($D$). The pale green shaded PC cycle (left side) is for all one's input condition, whereas the pale red shaded (right side) PC cycle is for all zero's input condition. The power-clock with $10\mu$m width has a +ve peak value close to 1.8V which increases to $1.825V$ for $100\mu$m and a $-$ve peak close to $-5mV$ at $10\mu$m which extends to $-50mV$ at $100\mu$m for all 1 input PC cycle. In general, tweaking the width regulates the power-clock swing. However, higher swings lead to higher non-adiabatic losses.

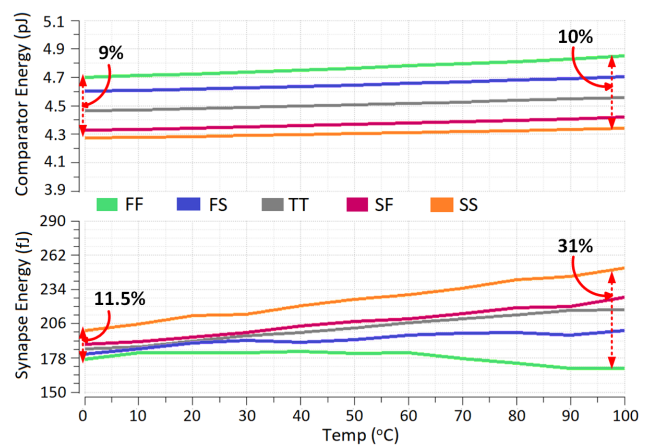

Fig. 18. Impact of process and temperature variation on the adiabatic capacitive synapse system for baseline scenario for five process corners. Top plot shows a constant energy/cycle variation across the temperature for memristor-based DLCC. Bottom plot shows smaller variation in the synapse energy at lower temperature which swells on increasing temperature.

Finally, to close this subsection we show results for a broader range of scenarios, where frequency significantly deviates from our baseline case (1MHz case is baseline scenario itself). After some coarse-grain optimisation on the balance between C and L, we report on the minimum discovered energy points at 4 different frequencies. For these tests we used a much simpler, but probably also much more representative method: we took the last sweeps of our $5\times$ input scenarios ($1\times$ unscrambled $+ 4\times$ scrambled as before) and averaged the energy dissipation across all scenarios to give an "indicative fair" value for the energy dissipation. We repeated this for the adiabatic and non-adiabatic designs. Results are shown in Table IV. For these exploratory scenarios we adopted two criteria for determining $C_E$: First, it should not be too large to reduce energy dissipation. Second, it should not be excessively low to reduce load variation. $L_{PC}$ was the calculated accordingly. None of the scenarios was aggressively optimised for energy. Results show that: a) significantly lower synaptic energy dissipation is achievable in a wide range of scenarios and b) similar energy figures are achieved in the range from 100kHz to 1MHz. Transistor channel lengths are minimum in all cases.

### C. Impact of Process Corners and Temperature

To measure the robustness of the proposed adiabatic capacitive synapse setup, we performed corner analysis, checking functionality and energy dissipation. For this entire section we use the baseline scenario with the cyclic input set (unscrambled). Frequency is in all cases reduced by 2.3% in order to bring the resonator frequency closer to its energy-optimum for the loading scheme used. Temperature (T) was varied from $0^oC$ to $100^oC$ for three "even" corners namely 'FF', 'TT' and 'SS' and two "skewed" corners that is 'FS' and 'SF'.

Fig. 18 show performance the five corners vs temperature. The adiabatic memristor-based $DLCC$ and synapse show a maximum variation of 10% and 31% respectively across all corners and 100 grad. A similar analysis was also performed on the non-adiabatic synapse as shown in Fig. 19. The synapse energy shows a small variation of 3% which is mainly due to the inverters that are used to charge and discharge the synaptic capacitance's. The DLCC dissipation, on the other hand varies by a max. of 11% across corners and T, which is slightly more than the adiabatic case.

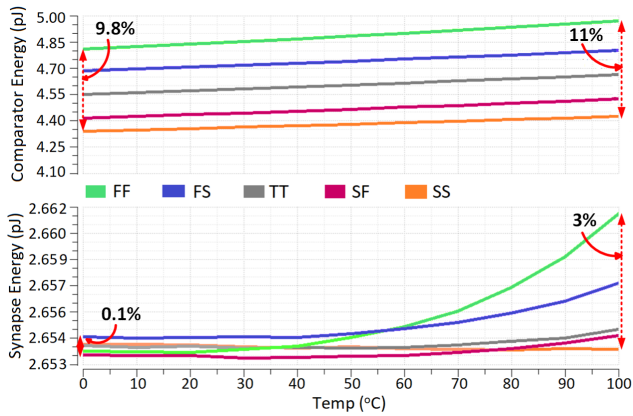

Fig. 19. Impact of process and temperature variation on the non-adiabatic capacitive synapse system for baseline scenario. Top plot shows a similar variation in the comparator energy as for the adiabatic system. Bottom plot shows much smaller variation in the synapse energy across corners and temperature, albeit atop a much higher baseline.

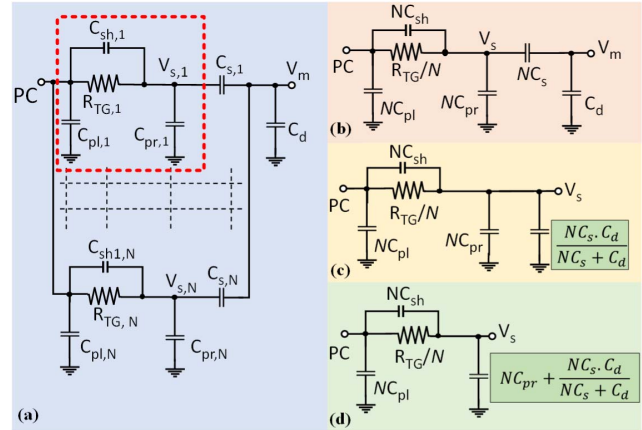

Fig. 20. (a) Equivalent RC network circuit of $N$-bit synaptic tree including ON-resistance ($R_{TG}$), shunt capacitance ($C_{sh}$) and parasitic capacitance's ($C_{pl}$ & $C_{pr}$) of the transmission gate (shown in red rectangle). (b), (c) and (d) shows how this circuit can be progressively simplified given that all indexed components are correspondingly equal, leading to the equation in (d). Notation as in Fig. 5.

From the figures we observe the following: First, the energy dissipation dependencies on corners are opposite for the synaptic tree and DLCC. The SS corner leads to maximum dissipation in the synapses but minimum in the DLCC. This is due to the dependence of synapse tree energy on its TGs' ON-resistance $R_{TG}$ (see eq. (7)) which, is a function of threshold voltage and, in turn, depends on the corner.

Next, although the comparator energy dissipation is different between adiabatic and non-adiabatic cases. The marginally lower energy dissipation in the adiabatic case is attributed to the fact that the adiabatic DLCC is subjected to a lower $V_{DD}$ on average while it is making its decision: with reference to Fig. 11 in the non-adiabatic case $V_{DD}$ is always invariably 1.8V whilst in the adiabatic case it has already dropped visibly by the time the output signals diverge. We note that the input from the synaptic tree is relatively stable and at its maximum when the DLCC is activated and at its most sensitive (when the positive feedback begins), so movement of the input is unlikely to be a contributing factor.

Finally, we notice practically flat temperature dependencies for the DLCC in both cases and for the synaptic set-up in the non-adiabatic test. In the adiabatic circuit case there is a more pronounced temperature dependence of energy dissipation, with most corners showing a slight increase in energy with temperature as might be expected by the increased resistances observed in the TGs gating each synapse. It is not clear exactly what causes the not strictly monotonic fluctuations observed on multiple corners.

### D. Loading Effect and Scaling

Here we examine the effect of increasing numbers of synapses on neuron energy dissipation and operating frequency. First the equivalent RC network for $N$-synapse/neuron is modelled as shown in Fig. 20. Total effective capacitance at node $PC$ and $V_m$ is approximated by:

$$C_{PC} = \begin{cases} C_E + NC_{pl,OFF} + NC_{sh} & \alpha = 0 \\ C_E + NC_{pl,ON} + NC_{pr,ON} + \\ \dfrac{NC_s \cdot C_d}{NC_s + C_d} & \alpha = 1 \end{cases} \quad (9)$$

$$C_T = NC_{pr} + \dfrac{NC_s \cdot C_d}{NC_s + C_d} \quad \alpha = 1 \quad (10)$$

where $\alpha \in [0, 1]$ denotes normalised synaptic load (how many synapses are active). The parasitic capacitance, $C_{pl}$ comprises of $2C_{gs,p} + C_{db,p} + C_{gs,n} + C_{sb,n}$ and is different for the ON ($C_{pl,ON}$) and OFF ($C_{pl,OFF}$) cases for the synapse switch. Similarly, $C_{pr}$ parasitic capacitance comprises of $C_{gd,p} + C_{db,p} + C_{gd,n} + C_{db,n}$. For $\alpha = 0$, all the synapses are switched OFF and node $V_s$ couples with the PC node via shunt capacitance. The shunt capacitance will be in series with the $C_T$ and since $C_T$ is very large (due to $C_s$ & $C_d$) the effective capacitance is only $C_{sh}$. Thus, the frequency at $\alpha = 0$ is close to its optimum. This can be visualise in Tables V & VI when all the synapse are OFF (0% loading) the frequency is close to the optimum.

In all scenarios below, the starting point is choosing the PC inductance $L_{PC}$ such that when all synapses are disconnected the nominal frequency ($f$) is $1\,MHz$. This is done for various choices of $C_E$ to illustrate the load variation penalty as the equalising capacitance is reduced.

Simulations are carried out on the baseline scenario (with the exception of the number of synapses) and variations on the LC combination as shown in Tables V and VI for 512x and 1024 synapses respectively. At 0% synapse loading we obtain shifts in optimum oscillating frequency of less than 1%, in both cases becoming very insensitive to changes in $C_E$ once that exceeds $\approx 200\,pF$. However, at 100% loading, the operating frequency decreases by up to a factor of 5x at $C_E$ value of $25\,pF$ (worst case: 1024 synapses). Results show how stabilising the frequency close to $1\,MHz$ requires larger

TABLE V
ENERGY CONSUMPTION/CYCLE FOR 512-SYNAPSE NEURON AT 0% AND 100% SYNAPTIC LOADING. f: OPTIMAL FREQUENCY OPERATION FOR EACH SCENARIO, $S_E$ AND $N_E$: ENERGY CONSUMPTION FOR THE ENTIRE SYNAPTIC TREE AND FOR A SINGLE NEURON SOMA (COMPARATOR) RESPECTIVELY

| $C_E$ (pF) | 0% Synapse loading | | | 100% Synapse loading | | |
|---|---|---|---|---|---|---|
| | f (KHz) | $S_E$ (pJ) | $N_E$ (pJ) | f (KHz) | $S_E$ (pJ) | $N_E$ (pJ) |
| 25 | 992 | 0.112 | 3.744 | 300.39 | 4.993 | 4.545 |
| 50 | 996 | 0.211 | 3.745 | 404.20 | 6.399 | 4.553 |
| 100 | 999 | 0.419 | 3.740 | 530.22 | 8.138 | 4.572 |
| 200 | 1000 | 0.833 | 3.738 | 662.69 | 10.351 | 4.577 |
| 500 | 1001 | 2.056 | 3.740 | 813.67 | 14.035 | 4.595 |
| 1000 | 1001 | 4.005 | 3.738 | 893.66 | 16.735 | 4.640 |

TABLE VI
SAME AS TABLE V, BUT FOR A 1024-SYNAPSE NEURON AND INCLUDING 50% SYNAPSE LOADING CASE

| $C_E$ (pF) | 0% Synapse loading | | | 50% Synapse loading | | | 100% Synapse loading | | |
|---|---|---|---|---|---|---|---|---|---|
| | $f(KHz)$ | $S_E(pJ)$ | $N_E$ (pJ) | $f(KHz)$ | $S_E$ (pJ) | $N_E$ (pJ) | $f(KHz)$ | $S_E$ (pJ) | $N_E$ (pJ) |
| 25 | 983.3 | 0.111 | 3.757 | 261.16 | 7.329 | 4.487 | 215.52 | 7.800 | 4.600 |
| 50 | 992.1 | 0.213 | 3.757 | 357.53 | 9.638 | 4.502 | 298.51 | 10.445 | 4.657 |
| 100 | 996 | 0.430 | 3.760 | 476.42 | 12.566 | 4.506 | 404.20 | 13.090 | 4.650 |
| 200 | 999 | 0.836 | 3.758 | 608.64 | 16.065 | 4.525 | 545.55 | 17.260 | 4.653 |
| 500 | 1000 | 2.085 | 3.767 | 772.50 | 20.755 | 4.529 | 704.23 | 22.440 | 4.650 |
| 1000 | 1001 | 4.200 | 3.755 | 865.05 | 25.025 | 4.550 | 814.33 | 26.500 | 4.650 |

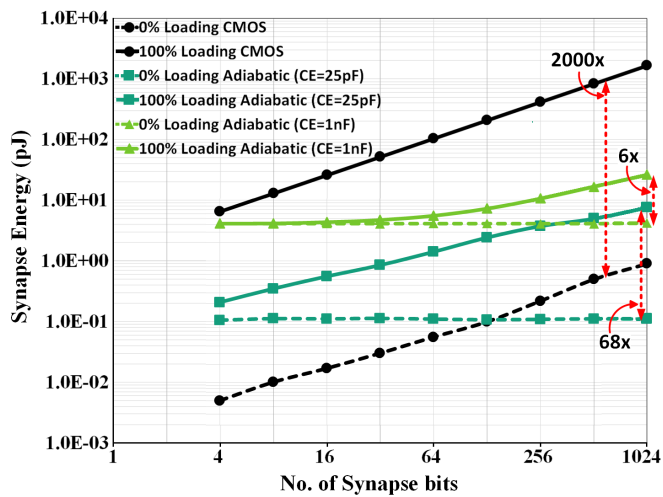

Fig. 21. Synapse tree energy for both adiabatic ($C_E = 25$pF & $C_E = 1$nF) and non-adiabatic designs. Solid traces: 100% synapse loading. Dashed traces: 0% loading. Substantial savings for the adiabatic case can be witnessed. Non-adiabatic shows 2000x times increment in energy for 100% loading in comparison to 0%, while adiabatic design for $C_E$ 25pF & 1nF shows 68x and only 6x increment in energy respectively for 1024 synapse.

## IV. DISCUSSION

A striking observation for tables V and VI is how in the adiabatic case even a fairly large number of synapses (e.g. 512) using a large equalising capacitor for reducing load variation of the optimum operating frequency still consumes energy comparable to its soma: ≈16.75 $pJ$ vs. ≈4.65 $pJ$ for a $1 nF$ equaliser cap. Notably, this decreases load variation of operating frequency to a bit over 10%. We see two implications of this: First, it suggests that extremely low-power neural networks may be attainable using adiabatic techniques. Second, it provides motivation for potentially "adiabaticising" or otherwise optimising the soma design as well for a further round of non-negligible performance gains, especially for neurons with lower synapses-to-soma convergence ratios.

Next, we note that load variation of operating frequency can reach extremely high levels whilst at the same time is coupled in a trade-off with the equaliser cap and consequently energy consumption. We foresee that the precise impact of load variation on performance will depend on two key factors: The first one is the sensitivity of energy dissipation on the discrepancy between operating and optimal frequency. Results from Figs. 14 and 15 reveal a dependence that we may prove quite steep, but a much broader range of frequencies around the optimum needs to be investigated to determine precisely how steep. The second factor is the average loading fluctuation over time for the neural network. It is possible to envisage that large and continuously ON neural networks may eventually achieve very stable loading regimes by virtue of large-scale averaging, possibly stable enough to allow a slow frequency tuner to adapt to the optimum over time (in the spirit of maximum power-point trackers in solar cells [41]). For instance; suppose a billion neurons are active in one cycle, then in the immediately next cycle chances are that the number of active neurons is not going to be dramatically different from a billion neurons. Importantly, this is a tendency and not equalising capacitance in exchange for an increase in energy consumption. We note that due to the sub-linearity of the synapse energy vs. total capacitance, whereby as total capacitance $C_T$ increases, so does oscillating period $T_{PC}$ -see eq. (7)-, we expect the 50% loading case to be much more similar to the 100% loading case, which indeed is borne out of the results in Table VI.

Fig. 21 shows the logarithmic energy plot for both adiabatic and non-adiabatic designs. In the non-adiabatic case synaptic tree energy scales linearly with number of synapses, reaching over $1.5 nJ$ for 1024 synapses. Adiabatic designs exhibit more than 90% energy saving for N synapse/neuron.

a hard fact: it becomes easier to "smooth out" the activity profile of the network, but not guaranteed. It remains to be seen what performances are practically achievable given this (very application- and architecture-dependent) interaction.

Another effect is the Kickback noise, a major source in DLCC occurring due to the coupling of regenerative nodes to the input of the comparator through the parasitic capacitance's of the transistors [42]. However, given the very large capacitive loading on one terminal ($V_m$ node) of the DLCC (which helps to suppress the thermal noise via the $KT/C$ mechanic) and the fact that the other terminal is driven by a voltage source, we can predict that the effect will remain quite limited. This is another imperfection we would like to treat in the dedicated "imperfections and non-idealities" publication together with mismatch and process variation effects.

We also foresee that the proposed system is actually arrange-able into crossbar configuration: presynaptic axons (inputs) map to horizontal lines as selector terminal ($SL$), the lines feeding $V_m$ map to vertical lines as the bit line ($BL$) and finally the PC terminal act as a word line ($WL$). The only difference is that, unlike memristive arrays, the capacitors will not create DC paths and can thus in principle compute without dissipating energy themselves; that is entirely a result of line resistances and other parasitic (whereas in memristive arrays the power dissipation is an essential aspect of the computation process). In terms of area, the metal-insulator-metal (MIM) capacitors used as synapses utilize approx. $22 \times 22$ $\mu m^2$ for $1pF$ fixed synapse capacitance value in the current technology used in this work. In contrast, memristive devices (before electroforming, when they effectively act as varicaps) have shown capacitance's of 18fF/sq. micron due to their very thin films (internal group data). In addition, using memcapacitors [15]- [18] would be beneficial in providing learning capability as weights can be adjusted by applying difference voltage across their terminals much as memristive devices do.

To add to the discussion on the impact of the process and temperature on the memristor model [38] and the $DLCC$ offset. The process corners can be -in principle- calibrated out. Temperature, on the other hand, is not automatically calibrated out and will affect the absolute values of the resistive states of the memristive devices as per [39]. However, because this is a differential design, some degree of automatic compensation is expected as both devices will be affected in the same way. The final effect on offset strongly depends on the operating regime of the memristive devices. In general, the closer together in resistance the devices are, the stronger the compensation effect of the differential topology (so 2× devices at 100k and 101k should in general experience far less offset dependence on temperature than a pair at 1k and 1Meg). This requires dedicated analysis and is the scope of our future work.

Finally, we note that this work has deliberately restricted itself to using a controllable, but realistic simulation set-up to uncover the design trade-offs involved in the engineering of the ACAN and provide sufficiently solid preliminary figures to determine how promising the avenue of adiabatic neurons is in the future (at least on the basis of the proposed design). The logical next steps would be to carefully study the effects of noise, imperfections and non-idealities along with the post-layout parasitics on a particular model architecture, however, that falls beyond the scope of the present study.

## V. SUMMARY AND CONCLUSION

This work demonstrates the amalgamation of energy recovery logic with an elegant capacitive synapse design [11] to reduce energy consumption, exposes the underpinning design trade-offs and provides preliminary data on expected performance. Our proposed Adiabatic Capacitive Artificial Neuron shows energy saving of even over 90% at $1MHz$ in comparison to its non-adiabatic counterpart, the performance disparity improving in favour of the adiabatic design as the convergence ratio i.e synapses/neuron is increased. Furthermore, we show robust functionality across a wide temperature range and corners despite an increased sensitivity of energy consumption on these parameters (while always remaining firmly substantially lower than the non-adiabatic design). Overall, we conclude that ACAN-based designs are likely to become of significant interest to the community due to their energy performance as: a) design techniques to operate them efficiently improve and b) variable capacitance (potentially memcapacitance [16], [18]) elements become more mature and allow in-situ reconfiguration without the overheads of switch banks.

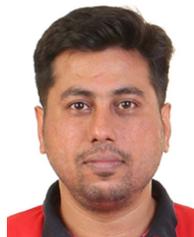

**Sachin Maheshwari** (Member, IEEE) received the bachelor's degree in electrical and electronic engineering from ICFAI University, India, the master's degree in microelectronics from the Birla Institute of Technology and Science, Pilani, India, and the Ph.D. degree in electronics engineering from the University of Westminster, London, U.K. He worked for two years as a Research Fellow at the University of Southampton, U.K. He is currently a Research Associate and a Post-Doctoral Researcher with the Centre for Electronics Frontiers, School of Engineering, The University of Edinburgh, Scotland, U.K. His research interests include artificial neural networks and neuromorphic computing, across energy recovery logic (adiabatic technique), and emerging technology (RRAM) for developing brain-inspired energy-efficient systems.

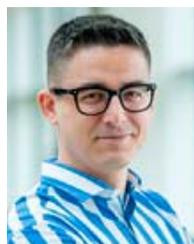

**Alexander Serb** (Senior Member, IEEE) received the degree in biomedical engineering and the Ph.D. degree in electrical and electronics engineering from Imperial College London, London, U.K., in 2009 and 2013, respectively. He is currently a Reader at The University of Edinburgh, Edinburgh, U.K. His research interests include cognitive computing, neuro-inspired engineering, algorithms and applications using RRAM, RRAM device modeling, and instrumentation design.

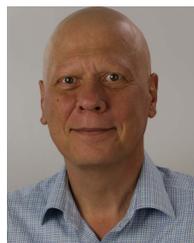

**Christos Papavassiliou** (Senior Member, IEEE) received the B.Sc. degree in physics from the Massachusetts Institute of Technology and the Ph.D. degree in applied physics from Yale University. He is currently working with the Department of Electrical Engineering, Imperial College London. He also works on memristor applications, sensor devices, and systems and antenna array technology. He has contributed to over 70 publications on weak localization, GaAs MMICs, and RFIC.

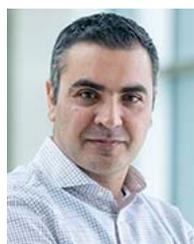

**Themistoklis Prodromakis** (Senior Member, IEEE) received the bachelor's degree in electrical and electronic engineering from the University of Lincoln, U.K., the M.Sc. degree in microelectronics and telecommunications from the University of Liverpool, U.K., and the Ph.D. degree in electrical and electronic engineering from Imperial College London, U.K. Then, he held a Corrigan Fellowship in nanoscale technology and science with the Centre for Bio-Inspired Technology, Imperial College London, U.K., and a Lindemann Trust Visiting Fellowship with the Department of Electrical Engineering and Computer Sciences, University of California, Berkeley. He holds the Regius Chair of engineering at The University of Edinburgh and the Director for the Centre for Electronics Frontiers. He also holds a Royal Academy of Engineering Chair in emerging technologies and a Royal Society Industry Fellowship. His background is in electron devices and nanofabrication techniques. His current research interests include memristive technologies for advanced computing architectures and biomedical applications. He is a fellow of the Royal Society of Chemistry, the British Computer Society, the IET, and the Institute of Physics. His contribution in memristive technologies was recognized as Blavatnik Award U.K. Honoree in physical sciences and engineering in 2021.